# Support Vector Machine Model for Currency Crisis Discrimination

Arindam Chaudhuri[1]

## Abstract

Support Vector Machine (SVM) is powerful classification technique based on the idea of structural risk minimization. Use of kernel function enables curse of dimensionality to be addressed. However, proper kernel function for certain problem is dependent on specific dataset and as such there is no good method on choice of kernel function. In this paper, SVM is used to build empirical models of currency crisis in Argentina. An estimation technique is developed by training model on real life data set which provides reasonably accurate model outputs and helps policy makers to identify situations in which currency crisis may happen. The third and fourth order polynomial kernel is generally best choice to achieve high generalization of classifier performance. SVM has high level of maturity with algorithms that are simple, easy to implement, tolerates curse of dimensionality and good empirical performance. The satisfactory results show that currency crisis situation is properly emulated using only small fraction of database and could be used as an evaluation tool as well as an early warning system. To the best of knowledge this is the first work on SVM approach for currency crisis evaluation of Argentina.

================================================================

1: Lecturer (Computer Science Engineering), Birla Institute of Technology Mesra, Patna Campus, Patna, Bihar, India. Email: arindam_chau@yahoo.co.in

# Support Vector Machine Model for Currency Crisis Discrimination

## I. Introduction

In past two decades, many economists have started to concede that self-fulfilling beliefs of investors have played a crucial role in emerging market financial crises. An important category of financial crisis is currency crisis prevailing in an economy. Timely identification and evaluation of currency crisis is a phenomenon of increasing interest to investors or creditors, financial institutions and governments alike to prevent impending failure in economy. Despite considerable progress on theoretical side, empirical models of currency crises have been shown to perform poorly (Berg et al., 1998; Goldfajn et al., 1999) and many economists and policy institutions have been struggling to develop adequate models to predict future financial crises (Kaminsky et al., 1997; Goldstein et al., 2000). Much of the empirical literature on currency crises however, still focuses on country specific macroeconomic factors that have ignored the development of an estimation algorithm by training model on data sets with reasonable accurate model outputs. Currency crisis is not exclusive to any specific economy. Globalization can feed the waves of economic distress across societies and national economies after the original economy witnesses its deleterious impact. Different countries are developing their own currency crisis discrimination models to deter disastrous consequences of ultimate financial distress. Identifying currency crisis using financial data is thus a traditional and modern topic of financial economics. The solution to this problem is an estimation technique by training a model on real life data defined into a binary set which provides reasonably accurate model outputs and helps policy makers to identify situations in which currency crisis may happen.

Currency crisis discrimination is basically binary classification problem and is identical to other classification problems such as text categorization (Joachims, 1998; Joachims et al., 2001), optical character recognition (Mori et al., 1992), intrusion detection (Mukkamala et al., 2002), speech recognition (Schmidt, 1996), handwritten recognition (Weston et al., 1999) etc. In binary classification, the problem has to learn to construct a number of separation boundaries or relations. Classification error rate is of great concern as there is error in determination of any one of decision boundaries or relations. Intuitively these methods can be interpreted as trying to construct conditional probability density for each class, than classifying by selecting the class with maximum aposteriori probability. For each data with high dimensional input space and very few samples per class, it is very difficult to construct accurate densities. In order to boost the performance of binary classifier Support Vector Machine (SVM) is used. SVM methods have reached a high level of maturity with algorithms that are simple, easy to implement, faster and with good performance. SVM originally designed for binary classification are based on Statistical Learning Theory developed by Vapnik (Corts et al., 1995; Vapnik, 1995). Larger and more complex classification problems have subsequently been solved with SVM. In designing Machine Learning algorithms, it is often easier to first devise algorithms to distinguish between two classes. SVM is learning machine that transforms training vectors into high dimensional feature space, labeling each vector by its class. It classifies data by determining set of support vectors, which are members of set of training inputs that outlines hyperplane in feature space (Vapnik, 1995). It is based on the idea of Structural Risk Minimization, which minimizes generalization error. The number of free parameters used in SVM depends on margin that separates data points and not on number of input features. SVM provides a generic technique to fit surface of hyperplane to data through the use of an appropriate kernel function. Use of kernel function enables the curse of dimensionality to be addressed and solution implicitly contains support vectors that provide a description of significant data for classification (Scholkopf et al., 2002). The most commonly used kernel functions are polynomial, gaussian and sigmoidal functions. Although in literature, the default choice of kernel function for most applications is gaussian. In training SVM we need to select kernel function and its parameters and value of margin parameter. The choice of kernel function and parameters to map dataset well in high dimension may depend on specific datasets. There is no method to determine how to choose an appropriate kernel function and its parameters for given dataset to achieve high generalization of classifier. The main modeling freedom consists in the choice of kernel function and corresponding kernel parameters which influences speed of convergence and quality of results. Furthermore, the choice of regularization parameter is vital to obtain good classification results.

The major prima face of this work is to develop an empirical model that helps policy makers to identify situations in which currency crisis may happen. Two alternative methodologies have been used for early currency crisis detection. The first approach is based on multivariate logit or probit model (Alvarez et al., 2000; Herrera et al., 1999; Kaminsky et al., 1998) and second approach used a warning based system viz., signals approach. Essentially it considers the evolution of a number of economic indicators tend to systematically behave differently prior to a crisis. If an indicator exceeds a specific threshold value, it is interpreted as a warning signal where a crisis may take place in next months (Herrera et al., 1999; Kaminsky et al., 1998). Here, empirical models built by training a SVM are presented. SVM provides a new approach to two category classification problem i.e., crisis or non-crisis with clear connections to the underlying Statistical Learning Theory (Burges, 1998). The approach is applied to evaluate currency crisis in Argentina during period 1999 – 2002. Different kernels are tried and less complex kernel that completely separates training data set with best *NSR* is chosen. The performance indexes achieved using SVM model are better than those previously reported. Overall, these results emphasize that only if we take into account systemic character of financial crisis it will improve our understanding and better prediction of occurrence of future crisis.

The paper is organized as follows. The concept of currency crisis is overviewed in section 2. The section 3 gives a brief review of Argentine economic currency crisis during 1999 – 2002. This is followed by a discussion of SVM in section 4. The next section presents SVM for currency crisis discrimination. A simulation example is given in section 6. Finally, the section 7 concludes the paper by outlining some general policy implications.

## II. Currency Crisis

Currency crisis is a situation in which an attack on the currency leads to sharp depreciation of currency, a large decline in international reserves or combination of two (Kaminsky et al., 1998). A currency crisis is identified by behavior of an index of *Speculative Pressure* (*ISP*) defined as (Herrera et al., 1999):

$$ISP_t = \Delta_{\%} Exchange_{rate} + \Delta_{\%} Interest_{rates} - \Delta_{\%} International_{reserves} \quad (1)$$

where, all variables (expressed in monthly percentage changes) are standardized to have mean zero and unit variance. An increase in index due to variation on these variables, for example a loss of international reserves, reflects stronger selling pressure on domestic currency (Kaminsky et al., 1998). A crisis is defined as period with unusual *pressure* (Herrera et al., 1999) $ISP_t > \mu + k\sigma$, where $\mu$ is sample mean and $\sigma$ is standard deviation of $ISP$ series and $k \geq 1$. As discussed in (Alvarez et al., 2000), $k = 1.5$ is selected to detect crisis event, while $k = 1.0$ is used to detect financially fragile event. Thus, binary variable $Crisis_t$ is defined as:

$$Crisis_t = \begin{cases} 1, ISP_t > \mu + k\sigma \\ 0, otherwise \end{cases} \quad (2)$$

There is a wide set of variables that can be used to build a model to explain crisis. In general, choice of variables is dictated by theoretical considerations and by availability of information on monthly basis (Kaminsky et al., 1998). In the process of determining currency crisis in Argentina following variables are used:

(i) Real Domestic Credit

(ii) International Reserves

(iii) Inflation

(iv) Oil Prices (Brent)

(v) Index of Industrial Equity Prices

(vi) Exchange Rate

(vii) Exchange Rate over-valuation using Hodrick-Presscott decomposition approach

## III. Argentine Economic Crisis (1999 – 2002)

Argentine economic crisis was a financial situation that affected Argentina's economy in the period 1999 – 2002. Macro-economically speaking the critical period started with the decrease of real GDP in 1999 and ended in 2002 with return to GDP growth, but the origins of collapse of Argentina's economy and their effects on population can be found in action before. The difficult political situation in Argentina over the years resulted in number of significant economic problems whereby country's industries were severely affected and unemployment prevailed. In order to cope with the situation new currency was issued to stabilize the economy for which new loans were procured. The country borrowed funds from different international agencies and was unable to pay them in time. The situation further became worse and the state eventually became unable to pay the debt and inflation cropped up. This continued for years which ultimately resulted in currency crisis situation in Argentina (Argentine Economic Crisis, Wikipedia).

## IV. Support Vector Machine

SVM provide a novel approach to the two-category classification problem viz., crisis or a non-crisis (Burges, 1998). The method has been successfully applied to a number of applications ranging from particle identification, face identification and text categorization to engine detection, bioinformatics and database marketing. The approach is motivated by Statistical Learning Theory (Cristianini et al., 2000). SVM is an estimation algorithm i.e. learning machine based on (Burges, 1998) which has following three steps:

(i) Parameter estimation procedure i.e. *training* from a data set

(ii) Computation of the function value i.e. *testing*

(iii) Generalization accuracy i.e. *performance*

*Training* involves optimization of a convex cost function; hence there are no local minima to complicate the learning process. *Testing* is based on the model evaluation using the most informative patterns in the data i.e., support vectors. *Performance* is based on error rate determination as test set size grows to infinity (Campbell, 1998). Considering $X_t$ and $y_t$ as set of variables and corresponding crisis evaluation respectively, such that SVM can be applied we have the following binary classifier:

$$Crisis_t = \begin{cases} -1, & ISP_t > \mu + k\sigma \\ 1, & otherwise \end{cases} \quad (3)$$

**IV. A. Linear SVM**

Suppose a set of $N$ multi-input single output training data points as $\{(X_1, y_1), \dots, (X_N, y_N)\}$. Consider the following hyperplane:

$$H : y = <w \cdot X> - b = 0 \quad (4)$$

where, $w$ is normal to $H$, $b/\|w\|$ is perpendicular distance to the origin and $\|w\|$ is the euclidean norm of $w$. Two hyperplanes parallel to equation (4) are:

$$H_1 : y = <w \cdot X> - b = +1 \quad (5)$$

$$H_2 : y = <w \cdot X> - b = -1 \quad (6)$$

with the condition, that there are no data points between $H_1$ and $H_2$ as shown in Figure 1 (http://www.ics.uci.edu/~xge/svm).

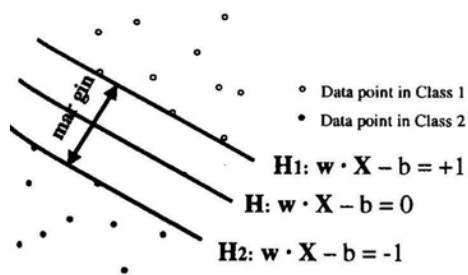

**Figure 1: Decision Hyperplanes generated by a linear SVM**

Let $d_+ (d_-)$ be the shortest distance from the separating hyperplane to the closest positive (negative) instance. The distance between $H_1$ and $H_2$ i.e. *the margin* is $(d_+ + d_-)$. It is easy to show that since $d_+ = d_- = 1/\|w\|$ then the margin is $2/\|w\|$ (Burges, 1998). Therefore, the problem to find a separating hyperplane with the maximum margin is achieved by solving the following optimization problem:

$$\min_{w,b} \frac{1}{2} w^T w \qquad (7)$$

subject to $y_i(<w \cdot X_i> - b) \geq 1$

The quantities $w$ and $b$ are parameters that control the function and are referred as the weight vector and bias (Cristianini et al., 2000). Equation (7) can be stated as convex quadratic programming problem in $(w,b)$. Using the Lagrangian formulation, the constraints will be replaced by constraints on Lagrange multipliers themselves. Additionally, a consequence of this reformulation is that training data will appear in the form of linear product between data vectors (Burges, 1998).

By introducing Lagrange multipliers $\alpha_1,........,\alpha_N \geq 0$, a Lagrangian function for the optimization problem can be defined as:

$$L_p(w,b,\alpha) = \frac{1}{2} w^T w - \sum_{i=1}^{N} (\alpha_i y_i (<w \cdot X_i> - b) - 1) \qquad (8)$$

To solve the optimization problem given in equation (8), one has to find the saddle point of equation i.e., to minimize equation (8) with respect to $w$ and $b$ and to maximize it over the non-negative Lagrange multipliers $\alpha_i \geq 0$ (Burges, 1998).

$$\frac{\partial L_p}{\partial w_i} = 0, i = 1,........,N \qquad (9)$$

$$\frac{\partial L_p}{\partial b} = 0 \qquad (10)$$

The gradients give the conditions (Burges, 1998):

$$w = \sum_{i=1}^{N} \alpha_i y_i X_i \qquad (11)$$

$$\sum_{i=1}^{N} \alpha_i y_i = 0 \qquad (12)$$

By substituting equations (11) and (12) into equation (8), one obtains Wolfe dual formulation (Burges, 1998; Campbell, 2000; Cristianini et al., 2000):

$$L_D(\alpha) = \sum_{i=1}^{N} \alpha_i - \frac{1}{2} \sum_{i,j=1}^{N} \alpha_i \alpha_j y_i y_j < X_i \cdot X_j > \qquad (13)$$

The notation has been changed from $L_p(w,b,\alpha)$ and $L_D(\alpha)$. Now to construct the optimal hyperplane one has to find the coefficients $\alpha_i$ that maximize equation (13) in the non-negative quadrant, $\alpha_i \geq 0$ under the constraint of equation (12). Solving for $\alpha_i$ and computing $b$ gives $w = \sum_i \alpha_i y_i X_i$.

Once SVM has been trained it is simple to determine on which side of decision boundary a given test pattern $X^*$ lies and assign the corresponding class label using $\mathrm{sgn}(< w \cdot X^* > -b)$. For the primal problem given by equation (8), Karush-Kuhn-Tucker (KKT) conditions are:

$$\alpha_i(y_i(< w \cdot x_i > +b) - 1) = 0, \forall i \qquad (14)$$

From condition in equation (14) it follows that non-zero $\alpha_i$ correspond only to vector $X_i$ that satisfy the equality,

$$y_i(< w \cdot x_i > -b) - 1 = 0 \qquad (15)$$

When the maximal margin hyperplane is found, only those points which lie closest to the hyperplane $\alpha_i > 0$ and those points are *support vectors* i.e., critical elements of training set. All other points have $\alpha_i = 0$. This means that if all other training points are removed and training is repeated, the same separating hyperplane would be found (Burges, 1998). In Figure 2, the points $a, b, c, d$ and $e$ are examples of support vectors (http://www.ics.uci.edu/~xge/svm).

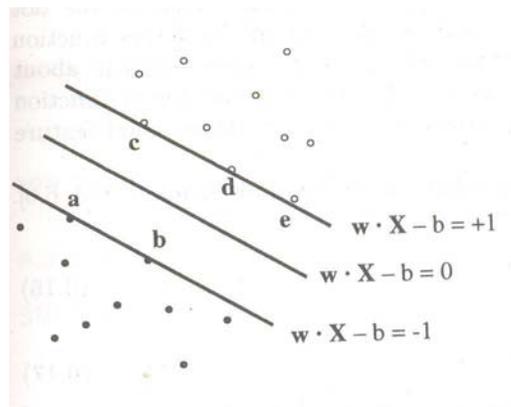

**Figure 2: Examples of Support Vectors**

Small problems can be solved by any general purpose optimization package that solves linearly constrained convex quadratic programs. For larger problems, a range of existing techniques can be used (Cristianini et al., 2000). The basic steps are as follows (Burges, 1998):

(i) Note the optimality KKT conditions which the solutions must satisfy

(ii) Define a strategy for approaching optimality

(iii) Decide on a decomposition algorithm so that only a portion of training data needs to be handled at a given time

**IV.B. Nonlinear SVM**

If the surface separating two classes is not clear, the data points can be transformed to a high dimensional feature space where the problem is linearly separable. Figure 3 denotes such transformation (Burges, 1998).

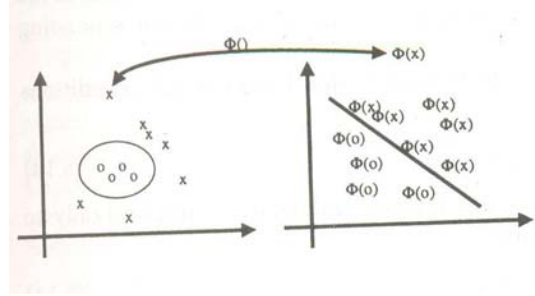

**Figure 3: A non-linear separating region transformed into a linear one**

Let the transformation be $\Phi(\cdot)$. The Lagrangian function in high dimensional feature space is given by,

$$L_D(\alpha) = \sum_{i=1}^{N}\alpha_i - \frac{1}{2}\sum_{i,j=1}^{N}\alpha_i\alpha_j y_i y_j <\Phi(X_i)\cdot\Phi(X_j)> \qquad (16)$$

Suppose that $<\Phi(X_i)\cdot\Phi(X_j)> = k(X_i, X_j)$ i.e., the dot product in high dimensional feature space defines a kernel function of input space. It is therefore necessary to be explicit about transformation $\Phi(\cdot)$ as long as it is known that kernel function $k(X_i, X_j)$ corresponds to a dot product in some high dimensional feature space (Burges, 1998; Campbell, 2000; Cristianini et al., 2000; Schölkopf, 2000). There are many kernel functions that can be used, for example (Burges, 1998; Cristianini et al., 2000) gaussian radial basis function kernel $k(X_i, X_j) = e^{(\frac{-\|x_i - x_j\|^2}{2\sigma^2})}$, polynomial kernel $k(X_i, X_j) = (<X_i \cdot X_j> + m)^p$.

The characterization of kernel function $k(X,Y)$ is done by means of Mercer's theorem (Burges, 1998; Cristianini et al., 2000). It establishes a positive condition that in case of finite subset of data points means that corresponding matrix of function $k(X,Y)$ has to be positive semi-definite. With suitable kernel, SVM can separate in feature space the data that in original input space was non-separable. This property means that we can obtain non-linear algorithms by using proven methods to handle linearly separable data sets (Campbell, 2000).

**IV.C. Imperfect Separation**

SVM can be extended to allow for imperfect separation (Burges, 1998; Cristianini et al., 2000) i.e., data between $H_1$ and $H_2$ can be penalized. The penalty $C$ is finite. Introducing non-negative slack variables $\xi_i \geq 0$ so that,

$$(<w \cdot X_i> -b) \geq 1 - \xi_i \text{ for } y_i = +1 \quad (17)$$

$$(<w \cdot X_i> -b) \leq 1 + \xi_i \text{ for } y_i = -1 \quad (18)$$

Adding to objective function a penalizing term the problem can now be formulated as:

$$\min_{w,b,\xi} \frac{1}{2} w^T w + C \sum_{i=1}^{N} \xi_i \quad (19)$$

$$\text{subject to } y_i(<w \cdot X_i> -b) + \xi_i \geq 0$$

$$\xi_i \geq 0$$

Using the Lagrange multipliers and the Wolfe dual formulation, the problem is transformed into,

$$\max_{\alpha} L_D(\alpha) = \sum_{i=1}^{N} \alpha_i - \frac{1}{2} \sum_{i,j=1}^{N} \alpha_i \alpha_j y_i y_j <\Phi(X_i) \cdot \Phi(X_j)> \quad (20)$$

$$\text{subject to } 0 \leq \alpha_i \leq C$$

$$\sum_i \alpha_i y_i = 0$$

The only difference from perfectly separating case is that $\alpha_i$ are now bounded above by $C$.

## V. SVM for Currency Crisis Discrimination

In the quest of determining appreciable results for currency crisis discrimination the following properties of SVM are utilized:

(i) Both training and test functions depend on data and kernel function. Even the need to evaluate dot product would result in less complexity of computing kernel. Thus, SVM circumvents both forms of curse of dimensionality; proliferation of parameters causing intractable complexity and over fitting.

(ii) Training algorithms may take advantage of parallel processing in several ways such as, evaluation of kernel and sum are highly parallelizable procedures.

(iii) SVM usually exhibit good generalization performance.

(iv) The choice of kernel is limitation of SVM approach. Some work has been done on limiting kernels using prior knowledge (Campbell, 2000). However, it has been noticed that when different kernel functions are used in SVM, they empirically lead to very similar classification accuracy (Campbell, 2000). In these cases, SVM with lower complexity should be selected.

(v) Although size of quadratic programming problem scales with number of support vectors, decomposition algorithms have been proposed to avoid problems with large data sets. For example, Sequential Minimal Optimization technique (Corts, Vapnik, 1995) explores an extreme case, where decomposition is made into sub-problems that are so small that an analytical solution can be found.

(vi) The performance of binary classifier is measured during test phase using its sensitivity, specificity and accuracy (Veropoulos et al., 1999) given by,

$$sensitivity = \frac{TP}{TP+FN}; specificity = \frac{TN}{TN+FP}; accuracy = \frac{TP+TN}{TP+TN+FP+FN};$$

where, *TP* is number of true positive classified cases (SVM correctly classifies); *TN* is number of true negative classified cases (SVM correctly classifies); *FP* is number of false positive classified cases (SVM labels case as positive while it is negative); *FN* is number of false negative classified cases (SVM labels case as negative while it is positive). For crisis discrimination, *sensitivity* gives percentage of correctly classified non-crisis and *specificity* percentage of correctly classified crisis events. These indexes can be easily converted to Type I and II errors as defined by *Kaminsky et al* (Kaminsky et al., 1998).

## VI. Simulation Example

In this section, SVM approach is applied to evaluate currency crisis in Argentina during the period 1999 – 2002. A monthly database with 232 observations (Alvarez et al., 2000) for each variable mentioned in section 2 is used. During this period 15 crises were detected, using *ISP* and $k = 1.5$. Two SVM models are developed to achieve the following:

(i) Classify any vector $X^*$ as representing a crisis or non-crisis event.

(ii) Predict from vector $X^*$ corresponding to specific month, if next month can be classified as a crisis or non-crisis event.

In first case, first 216 observations were used as the training set, of which 203 correspond to non-crisis and 13 to crisis event. In second case, SVM was trained with 215 observations, (202 for non-crisis and 13 for crisis event): for each vector $X_t$ corresponding $ISP_{t+1}$ was used. As given in (Herrera et al., 1999) and (Kaminsky et al., 1998), the information about sensitivity and specificity can be combined into a measure of noisiness of indexes viz., Noise/Signal Ratio ($NSR$) given by,

$$NSR = \frac{1 - sensitivity}{specificity}$$

This index measures the false signals as a ratio of good signals issued. The selection rule is to choose the model that minimizes $NSR$ (Herrera et al., 1999). Different kernels were tried and less complex kernel that completely separates training data set with best $NSR$ was chosen. In first case, a third order polynomial was selected; while a fourth order polynomial was the best kernel for second case. In both cases training $NSR$ was zero. During testing phase, $NSR$ was 0.143 and 0.153 for first and second cases respectively. All learning runs were performed on a 500 MHz Pentium IV machine using C++ program based on the Sequential Minimum Optimization Algorithm (Platt, 1999).

In order to compare SVM, results previously reported in similar studies based on the *signals* approach (Herrera et al., 1999) on multivariate probit model were (Alvarez et al., 2000) used. In both the studies, the data set corresponds to the period 1999 – 2002. Both studies considered the whole data set and did not use training or testing subsets. Additionally, they present results only for experiment 1. Table 1 presents SVM performance indexes evaluated using all observations. The performance indexes achieved using SVM model are better than those previously reported. The best $NSR$ using signal's approach is 0.19 while best $NSR$ using the present technique is 0.0458.

|  | Case 1 | Case 2 |
|---|---|---|
| **Polynomial Order** | 3 | 4 |
| **Number of Support Vectors** | 28 | 39 |
| *Sensitivity* (%) | 96.31 | 97.68 |
| *Specificity* (%) | 93.33 | 86.67 |
| *Accuracy* (%) | 96.12 | 96.97 |
| *NSR* | 0.0394 | 0.0257 |

**Table 1: SVM Performance Indexes**

## VII. Conclusion

This work presents a novel approach to evaluate Argentine currency crisis based on SVM. Experimental results show that third and fourth order polynomial kernel is generally best choice to achieve high generalization of classifier performance though it is often the default choice. The excellent results obtained through datasets show that currency crisis are properly emulated using only small fraction of database and could be used as an evaluation tool as well as an early warning system. The dependency of classifier accuracy can be exhibited on different kernel functions using different datasets. The classification accuracy may be further improved through different optimal value of parameter $C$. It will be interesting and practically more useful to determine some method for determining kernel function and its parameters based on statistical properties of given data. Then the proposed method can be effectively applied to other currency crisis datasets of other economies. Better results can be obtained if SVM classifier is integrated with the Fuzzy or Rough membership functions.